\documentclass[conference]{IEEEtran}
\IEEEoverridecommandlockouts
\usepackage{cite}
\usepackage{amsmath,amssymb,amsfonts}
\usepackage{algorithmic}
\usepackage{graphicx}
\usepackage{textcomp}
\usepackage{xcolor}
\usepackage{url}
\def\BibTeX{{\rm B\kern-.05em{\sc i\kern-.025em b}\kern-.08em
    T\kern-.1667em\lower.7ex\hbox{E}\kern-.125emX}}
\begin{document}

\title{Centimeter Positioning Accuracy using AI/ML for 6G Applications}
\author{\IEEEauthorblockN{Sai Prasanth Kotturi\IEEEauthorrefmark{1}, Radha Krishna Ganti \IEEEauthorrefmark{2}}
\IEEEauthorblockA{Department of Electrical Engineering, 
Indian Institute of Technology Madras\\
Email: \IEEEauthorrefmark{1}ksaiprasanth7@smail.iitm.ac.in,  \IEEEauthorrefmark{2} rganti@ee.iitm.ac.in
\vspace{-10pt}
}
}

\maketitle
\begin{abstract}
    This research looks at using AI/ML to achieve centimeter-level user positioning in 6G applications such as the Industrial Internet of Things (IIoT). Initial results show that our AI/ML-based method can estimate user positions with an accuracy of 17 cm in an indoor factory environment. In this proposal, we highlight our approaches and future directions.  
\end{abstract}

\section{Introduction}
As we progress towards 6G, several stakeholders from industry, academia, and standards bodies have recognized user positioning as a fundamental technology~\cite{yin2022analysis}. Conventional positioning schemes are either Radio Access Technology (RAT)-dependent or RAT-independent~\cite{keating2019overview}. RAT-dependent schemes require signalling from cellular networks and can be further categorized into five types depending on the type of measurement they rely on. Time-based methods~\cite{behravan2022positioning} use Time of Arrival (ToA) and Time Difference of Arrival (TDoA) measurements. Angle-based methods~\cite{behravan2022positioning} use Angle of Departure (AoD) and Angle of Arrival (AoA) measurements. Hybrid methods use a combination of time and angle measurements. Power-based and carrier-phase-based methods also exist. On the other hand, RAT-independent methods require signalling from satellite systems such as the Global Navigation Satellite System (GNSS). Conventional methods, though adequate in outdoor and Line-of-Sight (LoS) dominant scenarios, fail to achieve the desired accuracy in indoor and Non-Line-of-Sight (NLoS) dominant scenarios. This is because Non-Line-of-Sight (NLoS) paths lead to incorrect determination of timing, angle, power, and phase estimates. 

The only way to correctly estimate parameters such as ToA, TDoA, and AoA in NLoS scenarios is through the knowledge of channel characteristics either in the form of Channel Impulse Response (CIR), Power Delay Profile (PDP), or Delay Profile (DP). Such parameter estimation involves highly non-linear mappings, which are best learned using AI/ML techniques. AI/ML models can also be trained to estimate the user position directly, bypassing angle and time estimates. Depending on the output of the AI/ML model, positioning methods are of two types. The first is AI/ML-assisted methods that use raw channel measurements to perform intermediate tasks like LoS/NLoS classification or ToA estimation. The output of the intermediate tasks can aid subsequent AI/ML or non-AI/ML approaches for position estimation. The second is Direct AI/ML methods, in which an AI/ML model takes raw channel measurements as input and predicts the position as the output. In our research, we first look at Direct AI/ML methods applied to one example of an NLoS dominant scenario, the Indoor Factory. We note that irrespective of the type of AI/ML method and channel scenario, AI/ML-based positioning could be performed at either the UE side or the Network (base station) side within the Location Management Function (LMF) of the core network.

\section{Indoor Positioning Enhancement Using Direct AI/ML Methods}
We formulate the UE position estimation as a supervised learning task in which we train a neural network to estimate the position. Our current neural network model is a ResNet-like architecture that takes multiple indoor CIR and Reference Signal Received Power (RSRP) estimates as input and returns as output, the predicted user position in the horizontal plane $(\hat{x}, \hat{y})$.

\subsection{Indoor Factory System Model}
\begin{figure}[h]
\vspace{-15pt}
    \centering
    \includegraphics[width=0.46\textwidth]{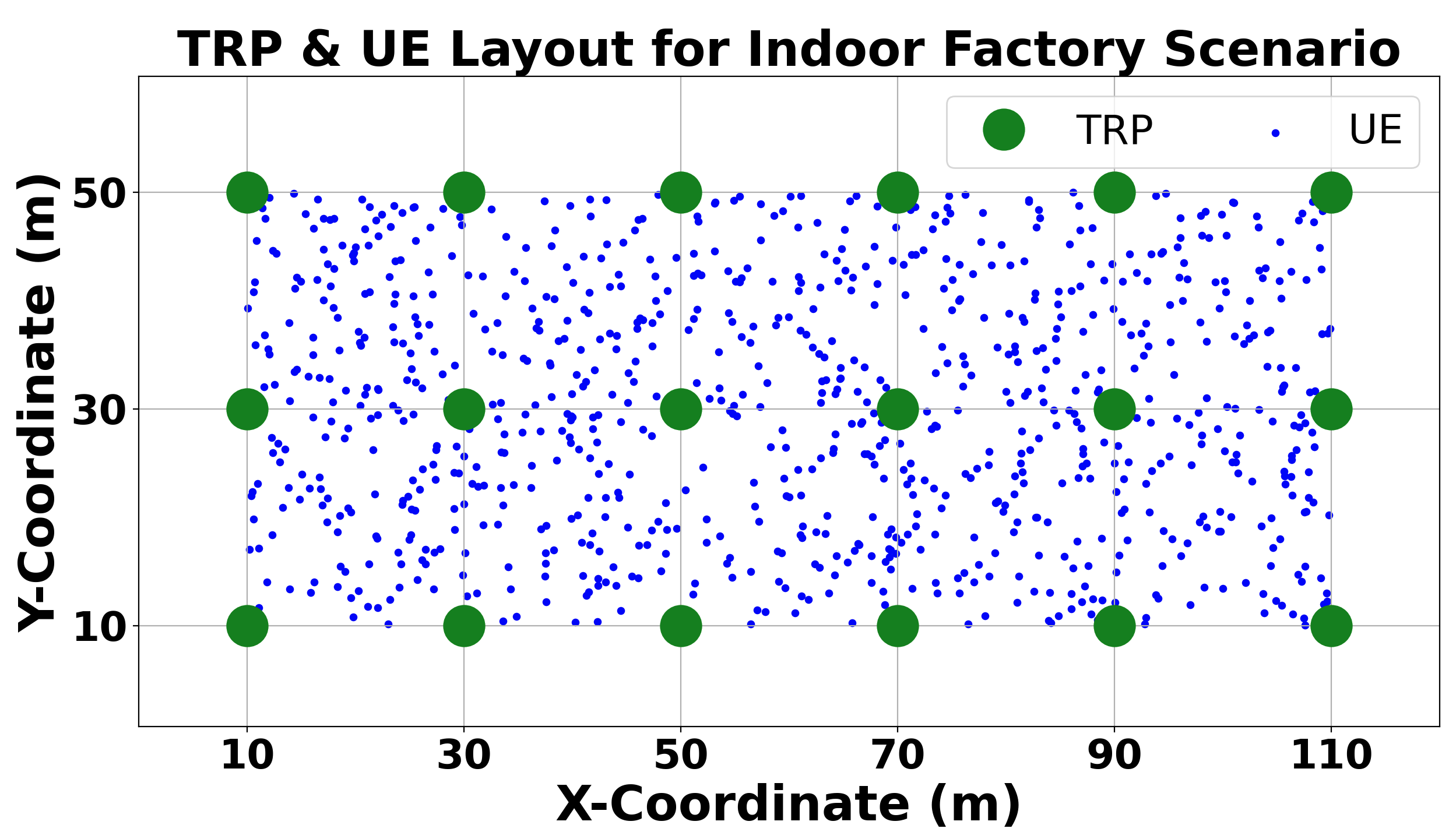}
    \vspace{-10pt}
    \caption{Indoor Factory Layout: Big dots represent fixed TRP locations, while smaller dots indicate uniformly distributed UE positions.}
    \label{fig: TRP and UE Layout for Indoor Factory Scenario}
    \vspace{-10pt}
\end{figure}

As an initial channel scenario (motivated by the Release 18 3GPP Study Item~\cite{3gpp_38_843}), we choose the InF-DH\cite{3gpp-38-901}, which stands for Indoor Factory with Dense Clutter and High Base Station Height. As illustrated in Fig.~\ref{fig: TRP and UE Layout for Indoor Factory Scenario}, the factory is a $120m \times 60m$ rectangular area with 18 Transmission Reception Points (TRPs) arranged in a $3\times6$ grid pattern. User Equipment (UEs) can be located with uniform randomness within the convex hull of the TRPs. The factory is also characterized by clutter objects and obstacles such as machines, robots, and walls. The clutter creates a highly NLoS channel scenario and is characterized by dimensions of $6m \times 2m$ with a $60\%$ clutter density.
\subsection{Proposed Approach}
Each UE has access to reference signal transmissions from up to 18 TRPs. From these reference signals, the UE can estimate the CIR and the RSRP. Other possible UE measurements include, but are not limited, to ToA and AoA. In our work, the training dataset~\cite{wireless_intelligence} consists of $80000$ instances of CIR and RSRP values obtained from downlink reference signal transmissions. The size of the CIR is $18 \times 256 \times 2$, where 18 is the number of TRPs, 256 is the CIR length (channel taps), and 2 corresponds to the real and imaginary parts of the CIR coefficients. For each instance of data, the UE also has 18 RSRP values. We have arranged the CIR and RSRP in an interleaved manner, resulting in dimensions of $36 \times 256 \times 2$. Each $256 \times 2$ CIR is sequentially followed by its associated RSRP value, which is replicated $256 \times 2$ times. This combination of CIR and RSRP works well for the following reasons: Since the model input is image-like, convolutional layers in the ResNet are well suited to learn the relevant CIR and RSRP features required to predict the position. Furthermore, the RSRP serves the purpose of aiding the model by assigning an implicit importance to these features.
\begin{figure}[h]
    \centering
    \vspace{-10pt}
    \includegraphics[width=0.43\textwidth]{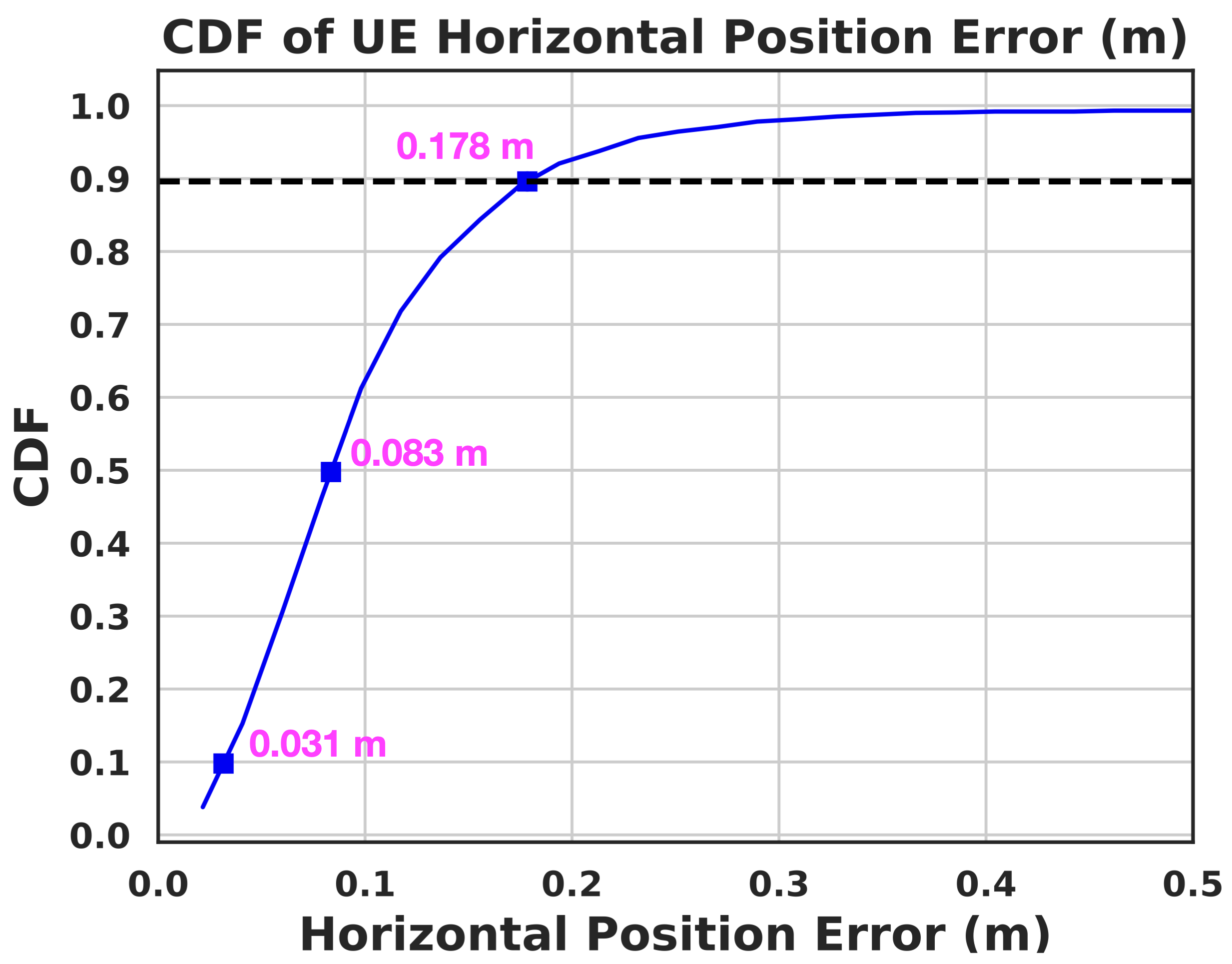}
    \vspace{-10pt}
    \caption{CDF of Inference Horizontal Positioning Error (m)}
    \label{fig: CDF of UE Horizontal Positioning Error for 18 TRPs}
    \vspace{-10pt}
\end{figure}
\subsection{Experimental setup} 
The input for our model is a preprocessed dataset, which is a combination of CIR and RSRP values. The AI/ML model training dataset consists of 80,000 data points divided into two subsets. The training set comprises $98\%$ of the original data, consisting of 78,400 data points. It includes the CIR and RSRP from all 18 TRPs as model inputs, with the corresponding ground truth labels (position) that serve as the model outputs. In the model training phase, we use 15 CNN layers that are trained with Adam optimiser and the batch size is set to 64. Initially, we set the Learning Rate (LR) to 2e-3 and decrease exponentially with a decay rate of 0.78 every 2 epochs. Once it reaches the minimum LR which is set at 9e-6, we increase the LR every 20 epochs till 1e-4 and then resume the decay process. This cycle continues until the model completes a total of 300 epochs. These changes are aimed at improving the accuracy of the model in estimating the position. For tuning hyperparameters and mitigating overfitting, a validation set is created, representing $20\%$ of the training data (15,680 data points). Finally, we consider the test dataset to be 1600 data points, which will be $2\%$ of the original dataset used for unbiased evaluation of model performance.

\subsection{Experimental Results}
Fig.~\ref{fig: CDF of UE Horizontal Positioning Error for 18 TRPs} shows the Cumulative Distribution Function (CDF) of the horizontal positioning error in meters across $1600$ users. We observe that the trained ResNet achieves a positioning accuracy as low as $17.8$ cm at the $90^{th}$ percentile. Our proposed model demonstrates better performance compared to the current machine learning models using direct AI/ML (ranging from $<1m$ to 5m) \cite{chatelier2023influence} and hybrid machine learning with denoising and inpainting methods (0.20m to 0.40m) \cite{zhao2022enabling} in the prediction of user position in indoor scenarios with dominant NLoS conditions. Additionally, in relevance to the above results we also refer our readers to \cite{3gpp_38_843}, where several companies have reported achieving positioning accuracies (ranging from 0.3m to 5m) in indoor environments using direct AI/ML method.

\section{Design Considerations for AI/ML-based Positioning}
This section highlights key design considerations for both direct AI/ML and AI/ML-assisted methods. In our research, we emphasize the crucial role of model input in determining performance by exploring alternative channel characteristics like PDP and DP to enhance feature extraction apart from CIR and RSRP. We will also investigate the impact of reduced input size by considering both decreased number of TRPs and channel taps. Additionally, we will try to experiment on ground truth labels where we account for measurement errors during data collection and the potential absence of ground truth values, which will transition the AI/ML models from supervised to semi-supervised learning. Finally, our approach involves training and testing on real-time base station data from the IIT Madras 5G Testbed which will enable us to assess real-world constraints such as signal attenuation, propagation loss, interference and multipath fading.

\bibliographystyle{IEEEtran}

\end{document}